\documentclass[runningheads]{llncs}

\usepackage[T1]{fontenc}
\usepackage{amsmath,amssymb,amsfonts,mathtools}
\usepackage{graphicx}
\usepackage{booktabs}
\usepackage{array}
\usepackage{xcolor}
\usepackage{microtype}
\usepackage[hidelinks]{hyperref}
\usepackage[most]{tcolorbox}
\usepackage[normalem]{ulem}

\DeclareMathOperator{\supp}{supp}
\newcommand{\R}{\mathbb{R}}
\newcommand{\E}{\mathbb{E}}
\newcommand{\KL}{\mathrm{KL}}

\begin{document}
\title{GEMSS: A Variational Method for Discovering Multiple Sparse Solutions in Classification and Regression Problems}
\titlerunning{GEMSS: Variational Discovery of Multiple Sparse Solutions}

\author{Kate\v{r}ina Henclov\'{a}\inst{1} \and
V\'{a}clav \v{S}m\'{i}dl\inst{2}}
\authorrunning{K. Henclov\'{a} and V. \v{S}m\'{i}dl}

\institute{Datamole, Prague, Czech Republic\\
\email{henclova.research@gmail.com}
\and
Faculty of Electrical Engineering, Czech Technical University in Prague,
Czech Republic\\
\email{smidlva1@fel.cvut.cz}}

\maketitle

\begin{abstract}
In underdetermined regression and classification problems, multiple feature subsets often yield equivalent predictive performance. In applied settings, especially with $n \ll p$, high dimension or collinearities, it is valuable to provide a domain expert with a menu of statistically plausible explanations, rather than one arbitrary solution. This creates the need for appropriate methods. 
We present Gaussian Ensemble for Multiple Sparse Solutions (GEMSS), a method that uses a single variational mixture to approximate the corresponding multimodal posterior. Its evidence lower bound contains a built-in repulsion between the mixture's components, enabling the model to simultaneously produce several distinct sparse solutions.
We evaluate GEMSS on a novel, reusable benchmark. The ground-truth solution set and its structure are known by construction and set-level recovery metrics are evaluated. 
GEMSS consistently outperforms dedicated multiplicity methods (Enumeration LASSO, ALFESE), two strong sampling baselines that approximate the same posterior (Randomized-LASSO ensemble, BB-SSL), and naive iterative masking.
As solutions' overlap increases, the gap widens and additional ensemble restarts cannot close it. Only ALFESE proves competitive.
Further, GEMSS is validated on real-world datasets, producing multiple distinct and highly predictive solutions: the practical goal that existing methods struggle to meet.
The open-source Python package \texttt{gemss} is available (\url{https://github.com/kat-er-ina/gemss}) and democratized through a free online application at \url{https://huggingface.co/spaces/kat-er-ina/gemss}.

\end{abstract}


\section{Introduction}
\label{sec:intro}

Selecting a small, interpretable set of features that explains a response is a recurring task in scientific data analysis, e.g., biomarker discovery in omics or mechanism identification in physical chemistry. In these settings, the number of features $p$ often greatly exceeds the number of samples $n$, or the features exhibit high correlation.

Two consequences follow. 
First, when relevant and irrelevant predictors are correlated, the irrepresentable condition fails. Thus, classical sparse selectors cannot recover a unique support~\cite{hastie2015statistical}. 
Second, and more importantly for practice, several distinct sparse feature subsets explain the data equally well. When a single-solution method produces its choice, it obscures alternative explanations.

This is the problem of discovering multiple sparse solutions. 
It appears in the literature under many names:
alternative feature selection~\cite{bach2024alternative}, 
predictive multiplicity~\cite{marx2020predictive} 
and the Rashomon effect~\cite{breiman2001statistical}, 
statistically equivalent signatures~\cite{tsamardinos2012discovering}, 
Markov-boundary multiplicity~\cite{statnikov2013algorithms}, 
and multimodal optimization in evolutionary computation~\cite{li2016seeking}. 

In many real-world applications, the primary objective shifts from pure prediction to knowledge discovery and generating actionable insights. Thus, our practical goal is to facilitate assisted discovery: present a domain expert with a menu of distinct, statistically competitive solutions. For example, when finding biomarkers of a disease, a scientist must adjudicate candidate sets based on biological specificity and practical utility.

One straightforward approach is to perturb the data or the objective and re-solve the resulting single-solution problem many times, collecting the diverse optima. One can choose the randomized LASSO, alternative feature selection (ALFESE)~\cite{bach2024alternative} or the Bayesian-bootstrap / posterior-bootstrap family~\cite{newton1994weighted,fong2019posterior,nie2022bayesian}. 

\textbf{GEMSS (Gaussian Ensemble for Multiple Sparse Solutions)}~\cite{henclova2021spms}, in contrast, is a single coherent model for problems with multiple valid explanations. It is a variational method that utilizes a probabilistic model whose posterior is explicitly multimodal in order to directly approximate the multiple solutions.

\textbf{This paper} comprises the following contributions. 
GEMSS (Section~\ref{sec:gemss}): a single variational-mixture model that returns several distinct sparse solutions from one fit.
A reusable benchmarking framework (Section~\ref{sec:benchmark}) to measure recovery of multiple feature sets. 
The controlled comparison (Section~\ref{sec:experiments}) tests GEMSS through six research questions.
Application to real-world datasets (Section~\ref{sec:realworld}) to demonstrate practical usability.

\textbf{Our open-source software} ships ready for use: the Python package \texttt{gemss}, a separate repository with the benchmarking framework, and a free online application to facilitate easy adoption by non-coders.

\section{Multiple sparse solutions as posterior approximation}
\label{sec:framework}

\subsection{The common target}
Consider a generalized linear model with response $y$, design $X\in\R^{n\times
p}$ and coefficients $\beta\in\R^p$, under a sparsity-inducing Spike-and-Slab
prior $\pi(\beta)$ that places mass on supports $S=\supp(\beta)$ of small size.
Write the posterior over coefficients (equivalently, over supports),
\begin{equation}
\label{eq:posterior}
p(\beta \mid X, y) \;\propto\; p(y \mid X, \beta)\,\pi(\beta).
\end{equation}
When several sparse supports explain $y$ equally well, the posterior \eqref{eq:posterior} is multimodal: 
each near-equivalent support is a mode. Discovering multiple sparse solutions is the task of characterizing the modes of \eqref{eq:posterior}, i.e. returning a set $\{S_1,\dots,S_m\}$ that covers the high-probability supports.

This is the common target that three different method families approximate by different means: a variational optimization, a principled posterior-bootstrap sampler, and a heuristic resampling ensemble.

\subsection{Variational approximation (GEMSS)}
\label{sec:gemss}
The multimodality of \eqref{eq:posterior} is induced by the prior. GEMSS's default is the Structured Spike-and-Slab~\cite{andersen2014bayesian}, which fixes the sparsity level at exactly $D$ nonzeros by mixing over the size-$D$ supports
$A\in\mathcal{A}$,
\begin{equation}
\pi(\beta)=\frac{1}{|\mathcal{A}|}\sum_{A\in\mathcal{A}}
\Big[\textstyle\prod_{j\in A}\pi_{\mathrm{slab}}(\beta_j)\prod_{j\notin A}\pi_{\mathrm{spike}}(\beta_j)\Big],
\end{equation}
with $\pi_{\mathrm{slab}}$ a wide Gaussian and $\pi_{\mathrm{spike}}$ a narrow one at zero; when $\binom{p}{D}$ is too large the supports are subsampled. 
Standard Spike-and-Slab and Student-$t$ priors are also available, with weaker control over
sparsity.

GEMSS approximates the resulting posterior by a parametric mixture of $K$ diagonal
Gaussians,
\begin{equation}
q(\beta)=\sum_{k=1}^{K}\alpha_k\,\mathcal{N}\!\big(\beta;\mu_k,\mathrm{diag}\,\sigma_k^2\big),
\qquad \sum_k\alpha_k=1,\ \alpha_k\ge 0,
\end{equation}
and maximizes the evidence lower bound
$\mathcal{L}=\E_q[\log p(y\mid X,\beta)]-\KL\!\big(q\,\|\,\pi\big)$ by stochastic
gradient descent (Adam), using implicit reparameterization for the
mixture~\cite{figurnov2018implicit}. 
Missingness is handled natively: the per-sample likelihood is evaluated over the observed coordinates only, with no imputation or row deletion. 

Each component $k$ contributes one candidate solution (support), which we read off as either the $D$ features of largest $|\mu_k|$ (in artificial benchmarks with fixed support size)  or as $z$-score-based outliers (in real-world problems, where  variable-size supports are desirable).
A fitted model also exposes per-component mixture weights $\alpha_k$, per-feature posterior variances, and the ELBO value.

Repulsion between modes is already present in the objective: the mixture entropy term inside $\KL$ contains a $\log\sum_k$ over components, so two components placed on the same mode raise $\log q$ at the samples and are penalized. This is what makes the mixture place its components on distinct modes rather than collapsing, and -- as the experiments show -- is the most plausible source of its edge in the high-overlap regime. The method is thus a single model: prior, likelihood, and diversity are one objective, optimized once.

An optional support-space penalty on the Jaccard similarity between solutions is available to let the
user directly control diversity as suitable to the given practical problem. 
The penalty's effect is governed by its scale: it engages only once $\lambda$ is comparable to the converged ELBO magnitude (typically $\lambda\sim10^3$). Smaller $\lambda$ values are inert; values that are too large override the likelihood, and recovery collapses.
Forcing solutions apart helps when they are truly disjoint but penalizes the correct shared core when they overlap. Hence, when used, $\lambda$ should be scaled to the ELBO and kept small unless the solutions are
expected to be disjoint.
In our experiments, we keep the penalty off by default ($\lambda{=}0$).

\subsection{Posterior-bootstrap sampling}
\label{sec:bbssl}
The Bayesian bootstrap~\cite{newton1994weighted} and its modern posterior-bootstrap form~\cite{fong2019posterior} re-weight the likelihood with random Dirichlet weights and re-optimize, yielding samples that approximate the posterior with consistency guarantees. \textbf{Bayesian-bootstrap Spike-and-Slab (BB-SSL)}~\cite{nie2022bayesian} applies this directly to the Spike-and-Slab model of \eqref{eq:posterior}: each draw re-optimizes the Spike-and-Slab objective under random likelihood weights, with a proven matching rate to the posterior. BB-SSL is thus a principled sampling approximation of exactly the same target as GEMSS, only reached by Monte Carlo over randomized optima rather than by variational optimization of a parametric $q$.

\subsection{Heuristic resampling ensemble}
\label{sec:ensemble}
In practice, the method a data scientist actually reaches for is cheaper and carries no posterior guarantee: a \textbf{Randomized-LASSO ensemble (RLE)}.
It resamples rows and reweights penalties~\cite{meinshausen2010stability}, re-solves a plain L1
problem each time, and clusters the supports; it has no contraction or matching-rate theory of its own. It can be read as a non-Bayesian cousin of the posterior bootstrap: the same perturb-and-re-solve loop with ad-hoc weights instead of Dirichlet ones.
We flag this only as a structural analogy, not a claim of equivalence: as Section~\ref{sec:experiments} shows, it does not behave like BB-SSL -- it is markedly stronger.

\subsection{The dedicated multiplicity methods}
\label{sec:dedicated}
The three methods above all aim at the posterior \eqref{eq:posterior}. Unlike them, the established multiplicity methods do not, and we include them as the prior art a practitioner would reach for. \textbf{Enumeration LASSO}~\cite{hara2017enumerate} returns the $k$ lowest-objective supports, which cluster around the single LASSO optimum -- one mechanism's neighborhood, not distinct mechanisms. \textbf{Masking} forces diversity by removing already-selected features and re-running a single-solution selector, so its alternatives are diverse by construction rather than data-driven. \textbf{ALFESE}~\cite{bach2024alternative} is a constrained-optimization wrapper that adapts conventional selectors to return multiple solutions. None targets the set of high-posterior supports.

\section{A benchmark with known multiple solutions}
\label{sec:benchmark}

Comparing set recovery requires having problems where the ground-truth set $\{S_1^\star,\dots\}$ is known. Predictive accuracy alone cannot tell whether a method found the right, meaningful or practical solution or merely a good predictor.

\subsection{Controlled-overlap generator}
We generate data from a rank-$r$ latent mechanism shared by several planted
supports, so that each planted support reconstructs the same signal and is hence
provably of equal predictive value (per-support $R^2\approx 0.99$ on the
noiseless signal). 
A single parameter, \texttt{overlap}, sets the size of the shared core of features common to all solutions, so the difficulty ranges from disjoint solutions (\texttt{overlap}$=0$) to heavily overlapping ones (large \texttt{overlap}).
Signal and noise features are matched in marginal variance so that a method cannot cheat on variance; Gaussian noise, missingness (seeded NaNs), and class imbalance are optional stressors. Binarizing the latent target yields the classification variant.

\subsection{Set-level metrics}

\textbf{We score support recovery, not prediction:} every planted support spans the same signal subspace, so all solutions are equally predictive by construction (verified on held-out data), and the same metrics serve regression and classification. Two complementary set-level metrics suffice. 

\textbf{Union-F1} is the $F_1$ between the pooled returned features $\bigcup_k S_k$ and the pooled planted features $\bigcup_j S_j^\star$. It addresses the question whether the relevant features are recovered at all, regardless of how they are partitioned into solutions. 

\textbf{Dissimilarity} $\mathrm{dissim}=\tfrac{2}{m(m-1)}\sum_{a<b}\big(1-\mathrm{Jaccard}(S_a,S_b)\big)\in[0,1]$ is the mean pairwise $1-$Jaccard over the returned supports. It measures how distinct they are: $0 =$ identical $=$ collapsed to one solution, $1 =$ pairwise disjoint. A method recovers multiple distinct, valid solutions only when union-F1 is high and dissimilarity is nonzero.
Note that dissimilarity is a descriptor, not a quality score: a high value can be forced (as masking does by construction). Therefore, on synthetic data we read it against the planted truth --- the true dissimilarity (the mean pairwise $1-$Jaccard of the planted supports) --- since the goal is to recover the correct overlap, not merely to maximize diversity. On real data (no ground truth) we read it alongside each solution's held-out predictive $F_1$.

\subsection{Fair-comparison protocol}
\label{sec:fairness}

To ensure fair method comparison, we hold three controls fixed:
\begin{enumerate}
\item[i]\textbf{Equal task information.} Every method returns the same number of final solutions $m$ and uses the same fixed per-solution sparsity $D$ (matching the planted sparsity on synthetic data). Internal capacity
is not constrained to $m$: GEMSS fits $K\ge m$ mixture components and reduces them to $m$ by clustering (the same reduction the ensemble applies to its restart cloud), and the ensemble draws as many restarts as its budget allows ($K$ and the restart count are per-method capacity knobs, tuned like any other hyperparameter).
\item[ii]\textbf{Equal tuning.} Each method's main hyperparameter is selected on the same footing -- GEMSS at its
documented recommended configuration, BB-SSL's spike/slab penalties swept on a grid, and the ensemble's LASSO penalty $\alpha$ swept likewise; we report the ensemble at a single fixed $\alpha$ and note where oracle per-problem tuning would change nothing.
\item[iii]\textbf{Implementation-agnostic budget.} The implementations span PyTorch, R and scikit-learn, so wall-clock is not comparable. We therefore express the ensemble's cost as its number of restarts (independent fits) and report recovery as a function of that count rather than seconds.
\end{enumerate}

\textbf{Tuning grids.} Every method's main hyperparameter(s) are swept and the best per condition is reported (Table~\ref{tab:tuning}); the reduction to $m$ is the same clustering for all, so none is rewarded for producing extra candidates.

\textbf{Details} of the benchmarking framework are documented in the dedicated repository \url{github.com/kat-er-ina/gemss_testing}.

\begin{table}[t]
\centering
{
\caption{Hyperparameter grids. When swept, the best per condition is reported. The solution count $m{=}3$ and per-solution sparsity $D{=}10$ are fixed for all methods.}
\label{tab:tuning}
\addtolength{\tabcolsep}{6.0pt}
\begin{tabular}{ll}
\toprule
method & tuned hyperparameter grid \\
\midrule
GEMSS                       & no. components $K\in\{3,6,12,18,24\}$ (clustered to $m$); \\
                            & diversity penalty $\lambda{=}0$ (off; ablated over $\{0,\dots,10^6\}$, Sec.~\ref{sec:gemss}); \\
                            & Spike-and-Slab prior: $\sigma^2_{\mathrm{spike}}{=}0.1$, $\sigma^2_{\mathrm{slab}}{=}100$, slab weight $= 0.9$ \\
RLE        & penalty $\alpha\in\{0.005,0.01,0.02,0.05\}$; \\
                            & restarts $\in\{50,100,300,3000,30000\}$ \\
BB-SSL                      & spike $\lambda_0\in\{5,10,30\}$, slab $\lambda_1\in\{0.1,0.5\}$ \\
ALFESE                      & selector $=$ MI; diversity $\tau\in\{0,0.25,0.5,0.75,1.0\}$ \\
EnumLASSO                   & $\rho\in\{0.01,0.02,0.05,0.1\}$ \\
Masking   & $\alpha\in\{0.01,0.05,0.1\}$ \\
\bottomrule
\end{tabular}
}
\end{table}

\section{Research questions and experiments}
\label{sec:experiments}

\subsection{RQ1: Which method best recovers the set of valid multiple solutions across varying degrees of feature overlap?}

\textbf{The claim under test.} The posterior-approximating methods (GEMSS, BB-SSL, RLE) inherently target the true solution set and should therefore recover overlapping solutions better (achieve higher union-F1) than methods that enumerate by objective value or force diversity. Furthermore, these methods should yield comparable recovery performance.

\textbf{The setup.} 
We evaluate support recovery on a continuous regression benchmark parameterized by structural overlap. There are $n{=}100$ samples, dimension $p{=}200$, $m{=}3$ planted supports, $D{=}10$ per-solution sparsity, results aggregated over $25$ random seeds. We compare posterior approximations (GEMSS, BB-SSL, RLE) and dedicated multiplicity baselines (EnumLASSO, ALFESE, Masking).
All methods undergo rigorous hyperparameter tuning to ensure fair evaluation. GEMSS is evaluated at its empirically optimal variational capacity ($K$) and clustered ($K \to m$) via agglomerative-Jaccard consensus to match the constraints on output cardinality. We quantify total feature-set recovery using union-$F_1$ and assess topological fidelity using support dissimilarity (mean pairwise $1-\mathrm{Jaccard}$), benchmarking the latter strictly against the ground-truth structure.

\begin{table}[t]
\centering
\caption{RQ1. Overlap sweep. Recovery metric union-F1 with $95\%$CI $\pm.02$--$.06$; bold = best recovery per overlap. Best configuration per method is shown. Read dissimilarity against the true dissimilarity row. 
The wall-clock runtime per fit reports statistics over hyperparameter grid and $25$ seeds on one CPU node. GEMSS plateaus at $\sim2000$ iterations and returns all $m$ solutions from one fit, whereas the ensemble/BB-SSL figures are for their full restart/sample budgets.}
\label{tab:rq1_fair}
\small
\setlength{\tabcolsep}{4pt}
\begin{tabular}{l cccc cccc | cr}
\toprule
 & \multicolumn{4}{c}{union-F1} & \multicolumn{4}{c|}{dissimilarity} & \multicolumn{2}{r} {runtime [s per fit]}\\
\cmidrule(lr){2-5}\cmidrule(lr){6-9}\cmidrule(lr){10-11}
overlap & $0$ & $2$ & $4$ & $8$ & $0$ & $2$ & $4$ & $8$ &  & median (min – max)\\
\midrule
GEMSS     & \textbf{.84} & \textbf{.89} & \textbf{.91} & \textbf{.90} & .83 & .82 & .80 & .59 & & 21.4 ($18 - 304$) \\
ALFESE    & .75 & .72 & .71 & .69 & .89 & .67 & .46 & .00 & & 0.5 ($0.4 - 1.0$)\\
Masking   & .69 & .64 & .62 & .54 & 1.0 & 1.0 & 1.0 & 1.0 & & 0.0 ($0.0 - 0.2$) \\
BB-SSL    & .51 & .53 & .56 & .61 & .98 & 1.0 & 1.0 & 1.0 & & 18.5 ($5 - 2893$) \\
RLE       & .50 & .51 & .54 & .53 & .90 & .87 & .85 & .84 & & 55.7 ($14 - 149$) \\
EnumLASSO & .52 & .51 & .50 & .48 & .43 & .47 & .47 & .39 & & 0.1 ($0.0- 8.3$) \\
\cmidrule(lr){1-9}
\emph{true dissimilarity} & \multicolumn{4}{c}{---} & 1.0 & .89 & .75 & .33 & & \\
\bottomrule
\end{tabular}
\end{table}

\textbf{The finding.} GEMSS outperforms dedicated methods. As detailed in Table~\ref{tab:rq1_fair}, GEMSS leads union-F1 at every overlap level, with its margin over the baselines growing as overlap increases (significant throughout by paired per-seed win-rate). However, posterior-approximating methods do not coincide, i.e. shared Bayesian pedigree does not guarantee comparable recovery, as evidenced by the subpar performance of BB-SSL and RLE. ALFESE is the second strongest method. EnumLASSO and masking consistently trail.

\textbf{Fidelity to true structure.} Dissimilarity must be evaluated as fidelity to the true structure rather than by its raw magnitude. Masking and BB-SSL artificially force disjoint supports (staying pinned at a maximal $1.0$ dissimilarity), completely ignoring the underlying truth. In contrast, GEMSS recovers the correct amount of overlap—its dissimilarity declines in tandem with the ground truth, making it the clear leader in the distinct-and-valid criterion.

\textbf{Wall-clock runtime.} 
A single GEMSS fit ($\sim\!21$\,s, converging by $\sim\!2000$ iterations of $6000$) returns all solutions at once and is $2.6\times$ faster than the randomized-LASSO ensemble at its $3000$-restart budget, and comparable to BB-SSL's median but far steadier (BB-SSL ranges up to $\sim\!2900$\,s). We report these figures without a speed claim, as the budgets are not directly comparable.

\subsection{RQ2: Which methods return solutions that are \emph{both} structurally distinct \emph{and} highly predictive?}

\textbf{The claim under test.} While alternative solutions must be both predictive and structurally distinct, most methods fail to fulfill both requirements.

\textbf{The setup.} 
We test with 2 real metabolomic datasets: diabetes biomarkers~\cite{MTBLS1} and a metabolomics cohort with PCOS (polycystic ovary syndrome) patients and controls with preterm birth or lack thereof as the target~\cite{MTBLS12968}. 
We measure the predictive quality of each returned solution and their pairwise dissimilarity. 
Here, $F_1$ is the predictive binary-classification $F_1$ of the best of $11$ end models scored under nested cross-validation (outer CV: stratified $5$-fold; inner CV: $5$-fold for tuning $l_1$ and $l_2$ regularization parameters, fixed parameter setting otherwise). Note that no ground-truth is available.
In GEMSS, per-solution sparsity $D{=}10$ is used for prior setup, while raw solutions are extracted from the components via outlier detection at $STD{=}2.5$ (for flexibility more suitable to practical use) and then clustered to $m{=}8$ arbitrary candidates, using the same agglomerative-Jaccard consensus as the synthetic comparison protocol.
The eight-candidate setting and outlier detection match Section \ref{sec:realworld}.
Two datasets from Section~\ref{sec:realworld} are not included here. For the Arabidopsis data, all feature selection models found solutions with predictive $F_1 = 1.0$ for multiple end models. The food science dataset was not analyzed due to data privacy limitations.

\begin{table}[t]
\centering
\caption{RQ2, real data  with 8 sought solutions. We report predictive, binary-classification $F_1$ score of the best of $11$ end models. Each $F_1$ score was computed from results aggregated over 5 outer CV folds. Reported average and range were taken over \#sol distinct solutions, as well as the solutions' dissimilarity.}
\label{tab:RQ2}
\small
\setlength{\tabcolsep}{4pt}
\begin{tabular}{l cccc | cccc}
\toprule
 & \multicolumn{4}{c|}{diabetes ($n{=}132$, $p{=}222$)} & \multicolumn{4}{c}{PCOS$\to$preterm ($n{=}149$, $p{=}487$)} \\
\cmidrule(lr){2-5}\cmidrule(lr){6-9}
method & {\scriptsize \#sol} & {\scriptsize avg $F_1$} & {\scriptsize $F_1$ min--max} & {\scriptsize dissim} & {\scriptsize \#sol} & {\scriptsize avg $F_1$} & {\scriptsize $F_1$ min--max} & {\scriptsize dissim} \\
\midrule
GEMSS        & $8$ & $.94$ & $.92$--$.95$ & $.74$ & $8$ & $.85$ & $.82$--$.88$ & $.71$ \\
RLE          & $8$ & $.93$ & $.92$--$.95$ & $.88$ & $8$ & $.80$ & $.74$--$.84$ & $.94$ \\
EnumLASSO    & $6$ & $.93$ & $.91$--$.94$ & $.24$ & $1$ & $.88$ & ---          & $.00$ \\
Masking      & $8$ & $.83$ & $.75$--$.95$ & $1.0$ & $8$ & $.73$ & $.68$--$.83$ & $1.0$ \\
ALFESE       & $8$ & $.83$ & $.77$--$.90$ & $1.0$ & $8$ & $.72$ & $.67$--$.81$ & $1.0$ \\
BB-SSL       & $8$ & $.83$ & $.77$--$.92$ & $.98$ & $8$ & $.67$ & $.45$--$.84$ & $1.0$ \\
\bottomrule
\end{tabular}
\end{table}

\textbf{The finding.} On these real datasets, multiple distinct feature sets have comparable predictive properties (Table~\ref{tab:RQ2}). With GEMSS and strong end models, the per-solution predictive $F_1$ sits at $0.94$ on Diabetes and at $0.85$ on PCOS$\to$preterm data (and at $1.0$ across the methods for Arabidopsis data). These findings directly confirm predictive multiplicity. 
Moreover, the solutions' dissimilarity becomes the discriminator when the predictive metrics do not suffice.
EnumLASSO fails by yielding near-identical solutions (returning only $6$ distinct solutions out of the requested $8$ on Diabetes, and only $1$ on PCOS$\to$preterm). Iterative masking forces distinct solutions by construction (dissimilarity $=1.0$). For ALFESE, solution dissimilarity is a mandatory, tunable hyperparameter. BB-SSL too discovers mostly (Diabetes) or entirely (PCOS$\to$preterm) disjoint solutions but its predictive scores are the lowest. 
RLE is the only real competitor for GEMSS w.r.t. both metrics, with GEMSS achieving overall better predictive-$F_1$ scores.

\subsection{RQ3: How do GEMSS and RLE compare w.r.t. compute budget?}
\label{sec:RQ3}

\textbf{The claim under test.} Given enough restart budget, can RLE match the performance of GEMSS, making the variational model unnecessarily complex?

\textbf{The setup.} 
Since wall clock is not fairly comparable, we vary the ensemble's number of restarts (computationally cheap independent fits) and read recovery against that implementation-agnostic budget, with GEMSS's level marked.

\begin{table}[t]
\centering
{
\caption{RQ3. RLE's recovery (union-F1, best $\alpha$ per budget) as a function of restart budget vs.\ GEMSS ($D{=}10$, $15$ seeds).}
\label{tab:RQ3}
\addtolength{\tabcolsep}{3.5pt}
\begin{tabular}{lcccc|c}
\toprule
 & $R{=}50$ & $R{=}300$ & $R{=}3000$ & $R{=}30000$ & GEMSS \\
\midrule
overlap $=0$ & .48\,$\pm$.03 & .51\,$\pm$.03 & .51\,$\pm$.04 & .51\,$\pm$.04 & \textbf{.84}\,$\pm$.02 \\
overlap $=4$ & .49\,$\pm$.03 & .52\,$\pm$.03 & .55\,$\pm$.04 & .52\,$\pm$.03 & \textbf{.91}\,$\pm$.04 \\
overlap $=8$ & .49\,$\pm$.03 & .56\,$\pm$.05 & .53\,$\pm$.03 & .55\,$\pm$.03 & \textbf{.90}\,$\pm$.04 \\
\bottomrule
\end{tabular}
}
\end{table}

\textbf{The finding.} RLE cannot reach the performance of GEMSS by increasing the number of restarts (Table~\ref{tab:RQ3}): under union-F1 its recovery plateaus by a few hundred restarts and a $100\times$ further budget does not move it. The gap to GEMSS is large at every overlap.

\subsection{RQ4: Does GEMSS keep the lead as dimensionality grows?}
\label{sec:RQ4}

\textbf{The claim under test.} GEMSS's recovery edge is a property of the problem, not of one $(n,p)$ point, so it should persist as $n\ll p$ deepens.

\textbf{The setup.} 
We now test against the relevant rivals: ALFESE (by RQ1) and RLE (by RQ2). 
We fix $D{=}10$ and grow both axes: $n\in\{30,50,100\}$ and $p\in\{1000,2000,5000\}$ for the continuous regression setup, at a low and a high overlap ($2$ and $6$). We use $15$ seeds and each method at its best tuned hyperparameter, w.r.t. union-F1.

\begin{table}[t]
\centering
{
\caption{RQ4. Recovery vs. dimensionality 
(union-F1, $D{=}10$, $15$ seeds).}
\label{tab:RQ4}
\addtolength{\tabcolsep}{3.5pt}
\begin{tabular}{lccc|ccc}
\toprule
 & \multicolumn{3}{c|}{overlap $= 2$} & \multicolumn{3}{c}{overlap $= 6$} \\
 \cmidrule(lr){2-4}\cmidrule(lr){5-7}
$(n,p)$ & GEMSS & RLE & ALFESE & GEMSS & RLE & ALFESE \\
\midrule
$(30,1000)$  & \textbf{.63}$\pm$.04 & .48$\pm$.05 & .57$\pm$.04 & \textbf{.68}$\pm$.06 & .43$\pm$.06 & .52$\pm$.07 \\
$(30,2000)$  & \textbf{.60}$\pm$.04 & .41$\pm$.05 & .52$\pm$.08 & \textbf{.64}$\pm$.05 & .40$\pm$.04 & .49$\pm$.09 \\
$(30,5000)$  & \textbf{.52}$\pm$.06 & .38$\pm$.04 & .50$\pm$.06 & .48$\pm$.09 & .38$\pm$.03 & \textbf{.50}$\pm$.09 \\
$(50,1000)$  & \textbf{.64}$\pm$.01 & .57$\pm$.04 & .61$\pm$.07 & \textbf{.72}$\pm$.02 & .54$\pm$.05 & .57$\pm$.08 \\
$(50,2000)$  & \textbf{.64}$\pm$.03 & .57$\pm$.05 & .58$\pm$.04 & \textbf{.75}$\pm$.04 & .59$\pm$.04 & .55$\pm$.05 \\
$(50,5000)$  & \textbf{.62}$\pm$.03 & .56$\pm$.03 & .55$\pm$.06 & \textbf{.66}$\pm$.07 & .55$\pm$.07 & .54$\pm$.09 \\
$(100,1000)$ & .63$\pm$.02 & .56$\pm$.05 & \textbf{.67}$\pm$.04 & \textbf{.76}$\pm$.02 & .59$\pm$.04 & .67$\pm$.04 \\
$(100,2000)$ & .64$\pm$.01 & .57$\pm$.04 & \textbf{.68}$\pm$.05 & \textbf{.75}$\pm$.01 & .61$\pm$.03 & .70$\pm$.01 \\
$(100,5000)$ & \textbf{.66}$\pm$.03 & .58$\pm$.04 & .63$\pm$.05 & \textbf{.75}$\pm$.02 & .64$\pm$.05 & .63$\pm$.05 \\
\bottomrule
\end{tabular}
}
\end{table}

\textbf{The finding.} GEMSS generally outperforms the other methods across the dimensionality sweep (Table~\ref{tab:RQ4}). It ranks first in $15$ of $18$ cells (mean rank $1.17$), beating both the ensemble and ALFESE at almost every problem, including the extreme $p{=}5000$ regimes.
Pushing further, the recovery lead holds through $p{=}5000$ but narrows to parity at $p{=}10\,000$ (GEMSS: $0.56$ vs.\ ALFESE: $0.55$), where all methods degrade.

\textbf{Compute.} A single GEMSS fit scales gently with dimensionality (median $27$\,s at $p{=}500$ $\to$ $174$\,s at $p{=}10\,000$) and stays well below the restart-based ensemble ($47$\,s $\to$ ${\sim}1000$\,s), so GEMSS's compute advantage over the ensemble widens with $p$ (nearly $6\times$ at $p{=}10\,000$), while still converging within $\sim\!3400$ iterations.

\subsection{RQ5: How does recovery degrade under noise and missingness?}
\label{sec:RQ5}

\textbf{The claim under test.} In stress tests, GEMSS degrades gracefully.

\textbf{The setup.} 
At $n{=}100$, $p{=}200$, $D{=}10$, $overlap{=}4$, $15$ seeds, continuous regression, we sweep feature noise and the missing-rate separately for GEMSS, the ensemble, and ALFESE. GEMSS marginalizes missing entries natively, the ensemble mean-imputes, ALFESE has no missing-data path.

\begin{table}[t]
\centering
{
\caption{RQ5. Recovery (union-F1, mean$\pm95\%$CI) and solution
dissimilarity under high noise and missing data.
($D{=}10$, \texttt{overlap} ${=}4$, $p{=}200$). All dissimilarity confidence intervals $\pm0.03$--$0.06$. $\dagger$~ALFESE cannot run under missingness.}
\label{tab:RQ5}
\footnotesize
\setlength{\tabcolsep}{3.5pt}
\begin{tabular}{ll ccc | ccc}
\toprule
 &  & \multicolumn{3}{c|}{union-F1} & \multicolumn{3}{c}{dissimilarity} \\
 \cmidrule(lr){3-5}\cmidrule(lr){6-8}
sweep & level & GEMSS & RLE & ALFESE & GEMSS & RLE & ALFESE \\
\midrule
noise $\sigma$   & clean      & \textbf{.89}\,$\pm$.06 & .53\,$\pm$.03 & .72\,$\pm$.05 & .79 & .84 & .46 \\
                 & $\times4$  & \textbf{.83}\,$\pm$.05 & .55\,$\pm$.02 & .73\,$\pm$.04 & .66 & .89 & .46 \\
                 & $\times10$ & \textbf{.77}\,$\pm$.03 & .56\,$\pm$.04 & .65\,$\pm$.05 & .55 & .85 & .46 \\
                 & $\times20$ & \textbf{.69}\,$\pm$.05 & .55\,$\pm$.04 & .54\,$\pm$.05 & .45 & .84 & .46 \\
\midrule
missing frac.    & $10\%$ & \textbf{.78}\,$\pm$.04 & .56\,$\pm$.05 & $\dagger$ & .59 & .88 & $\dagger$ \\
                 & $25\%$ & \textbf{.76}\,$\pm$.03 & .62\,$\pm$.04 & $\dagger$ & .46 & .82 & $\dagger$ \\
                 & $50\%$ & \textbf{.73}\,$\pm$.03 & .59\,$\pm$.04 & $\dagger$ & .37 & .84 & $\dagger$ \\
\bottomrule
\end{tabular}
}
\end{table}

\textbf{The finding.} 
GEMSS leads union-F1 at every noise and missingness level (Table~\ref{tab:RQ5}) and degrades gracefully. With growing stress and lesser information, GEMSS's solutions grow increasingly similar, gliding toward structural collapse while maintaining comparably high recovery metrics.
ALFESE is more competitive under noise but degrades quickly and cannot run at all under missingness.
GEMSS's native missing-data likelihood is an asset where competitors must impute or, like ALFESE, cannot run at all.


\subsection{RQ6: What is the optimal number of components in GEMSS?}

\textbf{The claim under test.} The optimal number of mixture components (variational capacity) required to accurately map the posterior geometry is best decoupled from the number of planted solutions. Moreover, it depends directly on their structural similarity.

\textbf{The setup.} 
We test GEMSS across a capacity sweep ($K \in \{3, 6, 12, 18, 24\}$) on the overlapping regression benchmark ($D{=}10$, 25 paired seeds). To isolate the effect of internal capacity from the final output size, the raw variational components from each fit are clustered to a fixed deployment budget of $m{=}3$ candidate solutions. Support recovery is measured via union-F1.

\textbf{The finding.} 
As shown in Table~\ref{tab:capacity}, GEMSS's optimal number of components $K$ depends on the problem's overlap structure. Disjoint solutions are better fitted with higher $K$ values, which provide the necessary capacity to resolve separated, orthogonal posterior modes. Conversely, a smaller $K$ is better for heavy overlap, effectively preventing the fragmentation of highly correlated structural supports.

\textbf{Hyperparameter setup without an oracle.}
$K$ can be easily chosen by optimizing the ELBO value, while union-F1 decreases only by $0.028$ ($0.864$ vs.\ $0.891$). Moreover, such setup still exceeds every competing method's oracle score (GEMSS: $0.849$--$0.864$; next best, ALFESE: $0.708$).

\begin{table}[t]
\caption{RQ6. The effect of GEMSS' number of components $K$ (clustered to $m{=}3$) on union-F1 (CIs $\pm0.03$--$0.06$).}
\label{tab:capacity}
\setlength{\tabcolsep}{3.5pt}
\begin{center}
\begin{tabular}{lccccc}
\toprule
&  \multicolumn{5}{c}{overlap} \\
\cmidrule(lr){2-6}
$K$ & $0$ & $2$ & $4$ & $6$ & $8$ \\
\midrule
3  & .78 & .82 & .82 & \textbf{.93} & .88 \\
6  & \textbf{.84} & .87 & .86 & .92 & \textbf{.90} \\
12 & \textbf{.84} & \textbf{.89} & \textbf{.91} & .89 & .87 \\
18 & \textbf{.84} & .85 & .88 & .92 & .86 \\
24 & .82 & .84 & .88 & .88 & .80 \\
\bottomrule
\end{tabular}
\end{center}
\end{table}

\section{Real-world case studies}
\label{sec:realworld}

The benchmark investigates recovery under artificial conditions. However, the point of the method is \emph{assisted discovery}: handing a domain expert a few distinct, statistically likely candidate solutions to be further assessed. 

We illustrate this on three binary classification datasets that showcase different practical regimes. The diabetes dataset serves as a validation for the primary use case of $n<p$ with moderate $p$ and $n$. The Arabidopsis dataset demonstrates usability for an extremely low sample size $n{=}16$. Both these biological datasets comprise metabolomics measurements of different types. In contrast, the food science dataset combines measurements with modeled and exact values, posing the challenge of highly collinear features and noisy labels and uneven class ratio in a common $n>p$ problem.

The practical deliverable is always a small set of distinct, predictive, expert-checkable explanations. We omit the PCOS$\to$preterm dataset (used in RQ2) because the original paper is not yet available for the verification of our results.

To simulate practical use, these problems were solved manually using the GEMSS Explorer no-code application. Each time, 8 candidate solutions were requested, using z-score outlier detection for retrieving the feature sets from the components. 
Each candidate's predictive quality was judged by using classic classification methods as end models, with nested cross-validation  (outer CV: stratified 5-fold for diabetes and food science, LOO for Arabidopsis; inner CV: 5-fold for tuning l1 and l2 regularization parameters; fixed parameter setting otherwise), and computing the predictive F1 score from results aggregated across the outer CV folds.
The end-model registry comprises $11$ common classifiers: l1- and l2-regularized logistic regression, elastic net, support vector machine, $3$-nearest neighbors, XGBoost, random forest, decision tree, naive Bayes, and linear and quadratic discriminant analysis.

The results are summarized in Table~\ref{tab:rw_summary}. Full reports with details on the setup, detailed results and preprocessed public datasets are available in our repositories.

\begin{table}[t]
	\centering
	\caption{\textbf{Summary of predictive performance on real-world data:} $3\times8$\ candidate solutions, each with varying number of features ($D$). The mean pairwise dissimilarity ($1-$Jaccard) across the 8 candidates proves candidates' diversity. For predictive metrics, only the best end models and their F1 scores (aggregated over the outer-CV folds) are listed. For the Arabidopsis dataset, 7--10 models achieved the perfect F1 score.}
	\label{tab:rw_summary}
	\addtolength{\tabcolsep}{4.5pt}
	\begin{tabular}{l lll|lll|lll}
		\toprule
		\textbf{} 
		& \multicolumn{3}{c}{Diabetes}
		& \multicolumn{3}{c}{Arabidopsis}
		& \multicolumn{3}{c}{Food science} \\
        \cmidrule(lr){2-4}\cmidrule(lr){5-7}\cmidrule(lr){8-10}
		\multicolumn{1}{c}{Solution}
		& \multicolumn{1}{c}{D} & \multicolumn{1}{c}{F1} & model 
		& \multicolumn{1}{c}{D} & \multicolumn{1}{c}{F1} & models 
		& \multicolumn{1}{c}{D} & \multicolumn{1}{c}{F1} & model 
		\\ \midrule
		Candidate 1  & 21 & .872 & SVM   & 2 & 1.0 & 9x  & 3 & .928 & DT \\
		Candidate 2  & 31 & .879 & SVM   & 2 & 1.0 & 9x  & 4 & .945 & KNN \\
		Candidate 3  & 26 & .916 & SVM   & 4 & 1.0 & 10x & 5 & .930 & XGB \\
		Candidate 4  & 30 & .916 & SVM   & 1 & 1.0 & 9x  & 2 & .928 & RF \\
		Candidate 5  & 24 & .909 & SVM   & 1 & 1.0 & 7x  & 3 & .918 & LASSO \\
		Candidate 6  & 23 & .870 & SVM   & 2 & 1.0 & 9x  & 3 & .927 & XGB \\
		Candidate 7  & 30 & .901 & LDA   & 2 & 1.0 & 9x  & 6 & .918 & KNN \\
		Candidate 8  & 25 & .925 & LASSO & 2 & 1.0 & 9x  & 5 & .925 & XGB \\
		\midrule
        \multicolumn{1}{l}{Dissimilarity}
		& \multicolumn{3}{c}{.91}
		& \multicolumn{3}{c}{.97}
		& \multicolumn{3}{c}{.92} \\
        \bottomrule
	\end{tabular}
\end{table}

\textbf{Diabetes biomarkers.} The MTBLS1 metabolomics study~\cite{MTBLS1,salek2006diabetes} ($n{=}132$, $p{=}222$, $\sim$$36\%$ minority class) asks which urinary metabolites separate type-2 diabetes from controls. 
GEMSS returns 8 distinct sets of biomarkers, each independently predictive under nested CV (per-solution $F_1{=}0.87 - 0.92$), demonstrating the existence of multiple statistically comparable candidate solutions.

\textbf{Arabidopsis thaliana genomics.} The MTBLS2 study~\cite{MTBLS2,bottcher2009arabidopsis} separates wild-type from knockout plants from only $16$ samples over $41$ identified analytes: a regime where single-solution selectors lock onto one arbitrary subset or fail due to sample scarcity. 
GEMSS recovers multiple distinct solutions, all individually predictive ($F_1{=}1.0$) and comprising 1-4 features. They include the mechanism reported by the original study and statistically credible alternatives.

On metabolomics datasets, GEMSS' returned candidate solutions overlap the discriminating metabolites reported by the original studies \cite{salek2006diabetes,bottcher2009arabidopsis} and more \cite{brial2021,zhou2009}. The results are consistent with established domain knowledge and support the claim that the multiple selections are scientifically meaningful.

\textbf{Food science.} On this proprietary dataset ($n{=}175$, $p{=}48$ strongly collinear variables, $\sim$$11\%$ minority class), the goal was both knowledge discovery and a high-recall minority model. GEMSS's solution sets reproduced findings from prior domain-informed exploration and surfaced valid extensions, subsequently confirmed by a subject-matter expert~\cite{gillies2025softmatter}.

\section{Discussion}
\label{sec:discussion}

\textbf{Why the edge is where it is} (a mechanism, not a theorem). The behavior follows from one quantity, the mixture entropy $-\E_q[\log q]$ in the ELBO, which is an explicit parameter-space repulsion: maximizing it pushes component means apart, the same principle as repulsive mixtures~\cite{petralia2012repulsive,xie2020bayesian} and SVGD~\cite{liu2016stein}.
Heavily overlapping supports are nearby posterior modes; independent refits plus post-hoc clustering (the ensemble) cannot resolve nearby modes, whereas the repulsion separates them by construction, so the advantage grows with overlap (RQ1, RQ3), and our coupled-vs-independent ablation isolates exactly this effect.
Strengthening the repulsion through a parameter-space prior or likelihood tempering~\cite{ueda1998deterministic,rose1998deterministic} is the natural next step; we flag it rather than claim it.

\textbf{Threats to validity.} The benchmark is synthetic by the necessity of knowing the multiple true solutions. We mitigate it by finding multiple well-performing solutions for real-world datasets. 
The ensemble is a strong but specific pipeline; we tuned its $\alpha$ and gave it up to $30{,}000$ restarts rather than treat it as a straw man (RQ3), and we deliberately report it at a fixed rather than oracle $\alpha$. The high-overlap GEMSS edge rests on a modest seed count and one generator family; we state it as a localized effect, not a law. Theory is used only to set expectations, not as a contribution.

\textbf{Nonlinear problems.} Even though GEMSS inherently uses a linear model, it still enables a practically usable approach to solving nonlinear problems through feature engineering: add nonlinear transformation and interaction terms to the feature space. While this leads to an explosion in dimensionality and collinearities that typically render standard methods useless, GEMSS is applicable.

\textbf{Hyperparameter setup.} Firstly, the provided implementation of GEMSS (the no-code online application) offers heuristic default values and a user-friendly guide for manual tuning. Secondly, the ELBO provides a natural and usable optimization criterion for programmatic tuning.

\section{Conclusion}
\label{sec:conclusion}

We presented GEMSS, a feature-selection method that returns several distinct sparse solutions from a single fit. It is especially suitable for underdetermined problems with high dimensions, strong correlations or extreme sample scarcity.

We also release a reusable benchmarking framework, where the set of solutions is known by construction and their overlap is tunable. 

On this benchmark, GEMSS consistently outperforms its 5 competitors. It beats the tuned randomized-LASSO ensemble at every overlap -- a lead that widens with overlap and that more restarts cannot close. Only the ALFESE-wrapped mutual-information filter can compete but only at low solution overlap, with abundant samples and no missing values.

On real-world data, GEMSS achieves the practical goal that existing methods struggle to balance: returning candidate solutions that are both structurally distinct and highly predictive.

GEMSS is available as an open-source PyPI package \emph{gemss} (documented repository \url{github.com/kat-er-ina/gemss} with additional materials). The full benchmarking code and preprocessed real-world datasets are released in repository \url{github.com/kat-er-ina/gemss_testing}.
To facilitate easy adoption, we provide a free online application at \url{huggingface.co/spaces/kat-er-ina/gemss}.

\section{Acknowledgments}
Our thanks go to Datamole's Data Science team for their support and especially to Marek Nevole for his expert software engineering input.
We acknowledge the use of closely supervised AI tools in all parts of this research.

\bibliographystyle{splncs04}
\bibliography{references}

\end{document}